# Mobile Robot Localization via Indoor Positioning System and Odometry Fusion

Muhammad Hafil Nugraha
*Research Centre for Smart Mechatronics*
*National Research and Innovation Agency*
Bandung, Indonesia
muha167@brin.go.id

Fauzi Abdul
*Electrical Engineering Department*
*Telkom University*
Bandung, Indonesia
fauziabdul@student.telkomuniversity.ac.id

Lastiko Bramantyo
*Mechanical Engineering Department*
*Institute Technology of Bandung*
Bandung, Indonesia
13121129@mahasiswa.itb.ac.id

Estiko Rijanto
*Research Centre for Smart Mechatronics*
*National Research and Innovation Agency*
Bandung, Indonesia
estiko.rijanto@brin.go.id

Roni Permana Saputra
*Research Centre for Smart Mechatronics*
*National Research and Innovation Agency*
Bandung, Indonesia
roni.permana.saputra@brin.go.id

Oka Mahendra
*Research Centre for Smart Mechatronics*
*National Research and Innovation Agency*
Bandung, Indonesia
oka.mahendra@brin.go.id

*Abstract*—Accurate localization is crucial for effectively operating mobile robots in indoor environments. This paper presents a comprehensive approach to mobile robot localization by integrating an ultrasound-based indoor positioning system (IPS) with wheel odometry data via sensor fusion techniques. The fusion methodology leverages the strengths of both IPS and wheel odometry, compensating for the individual limitations of each method. The Extended Kalman Filter (EKF) fusion method combines the data from the IPS sensors and the robot's wheel odometry, providing a robust and reliable localization solution. Extensive experiments in a controlled indoor environment reveal that the fusion-based localization system significantly enhances accuracy and precision compared to standalone systems. The results demonstrate significant improvements in trajectory tracking, with the EKF-based approach reducing errors associated with wheel slippage and sensor noise.

*Keywords—indoor localization, wheel odometry, indoor positioning system, sensor fusion, extended Kalman filter, autonomous mobile robot.*

## I. Introduction

Wheeled Mobile Robots (WMRs) are widely utilized across various domains, including smart home services and indoor navigation assistance. Effective localization plays a critical role in these applications, enabling WMRs to navigate accurately and autonomously within indoor spaces. Notable examples of work in indoor positioning for WMRs include the cost-effective navigation system developed by Irwansyah et al. and the ultrasonic positioning approach by Li et al., both aimed at enhancing accuracy and reliability in indoor environments [1],[2].

Localization is a crucial capability for mobile robots, allowing them to determine their precise position in both indoor and outdoor settings. This accurate positioning is fundamental for essential functions such as navigation, obstacle avoidance, and path planning. In indoor environments, where GPS signals are often weak or unavailable, achieving reliable localization is particularly challenging and requires alternative solutions to maintain effective performance.

In autonomous mobility, various approaches have been developed to enhance localization accuracy. For instance, Farina et al. have significantly contributed to this area by focusing on the localization of an autonomous wheelchair [3]. Their work combines wheel odometry, Inertial Measurement Units (IMU), and Lidar, which are integrated through sensor fusion techniques. This fusion-based approach leverages the strengths of each sensor type, compensating for individual limitations and providing a more robust and accurate localization solution. Such advancements are vital for developing and deploying mobile robots in complex and dynamic indoor environments, where precise localization is critical for ensuring safe and efficient operation.

To provide a comprehensive solution for mobile robot localization, this study integrates wheel odometry data with an ultrasound-based indoor positioning system (IPS) using sensor fusion based on the Extended Kalman Filter (EKF) technique. By leveraging the strengths of both sensors, this approach aims to deliver a more accurate and robust localization system, crucial for the effective operation of mobile robots in complex indoor environments.

The primary objective of this work is to design and implement a sensor fusion-based localization system that:

1. Combines data from wheel odometry and IPS sensors to provide accurate real-time position estimates of a mobile robot.

2. Employs an Extended Kalman Filter (EKF) to effectively integrate disparate data sources, accounting for their individual uncertainties and error characteristics.

3. Improves localization accuracy and robustness in indoor environments compared to standalone odometry and IPS systems.

## II. Related Work

The effectiveness of indoor mobile robot localization systems has been the subject of extensive research. Yatigul et al. [4] investigated such systems by deploying a mobile robot on a square test circuit and examining the trajectories produced by visual and wheel odometry sensors using the Root Mean Square Error (RMSE) method. Teng et al. [5] developed a wheeled robot research platform for indoor localization, employing Adaptive Monte Carlo Localization (AMCL) and Extended Kalman Filter (EKF) algorithms.

Their system integrates data from Lidar, IMU, and wheel odometry sensors, along with Ultra-Wide Band (UWB) technology.

Zhang [6] utilized sensor fusion methods based on EKF to combine UWB and wheel odometry readings, benefiting from UWB's high accuracy and the low noise of odometry data. Liu et al. [7] also applied UWB and wheel odometry fusion, focusing on mobile robot localization in building corridors. Their results demonstrated that combining UWB with wheel odometry achieved higher accuracy than using either sensor alone.

Another approach involves Ultrasound-based IPS. These systems, such as Marvelmind Robotics, use stationary and mobile beacons to determine the position of the mobile beacon by measuring the Time of Flight (TOF) of ultrasonic pulses and applying trilateration techniques [8]. Lobo [9] compared odometry and IMU readings with error correction using trilateration and EKF, finding that these methods improved localization accuracy. Jimenez [10] analysed the Marvelmind IPS product on a mobile robot platform, assessing its performance in a small indoor area with both moving and static objects. Additionally, Garrote [11] integrated the Marvelmind IPS solution with laser scan measurements to enhance localization accuracy. Jimenez [10] shows limitations in handling fast movements, leading to significant positioning errors, and struggles with accuracy in varied or dynamic environments, especially when line-of-sight is obstructed. Garrote [11] uses a laser-based Particle Filter with A-IPS but lacks proprioceptive data, resulting in high computational demands and inefficiency in real-time applications. The proposed method in this paper, combining odometry and A-IPS with an Extended Kalman Filter (EKF), overcomes these issues, offering a faster, more efficient, and accurate localization solution in complex, dynamic settings.

This paper proposes using the Extended Kalman Filter (EKF) for state estimation due to its balance of accuracy and efficiency in handling moderate non-linearities, as found in fusing odometry and indoor positioning data. While the standard Kalman Filter (KF) is optimal for linear systems, it struggles with non-linear dynamics. The EKF overcomes this by approximating non-linear behavior through Taylor expansion. The Unscented Kalman Filter (UKF) offers more precision for stronger non-linearities but demands greater computational resources. Particle Filter (PF), though capable of handling both non-linearities and non-Gaussian distributions, is highly computational and requires a large number of particles for stability [11]. EKF thus provides a suitable compromise, ensuring effective real-time performance.

## III. PROPOSED METHOD

### A. System Architecture

Fig. 1 illustrates the architecture of the WMR system used in this study. Fig. 2 is WMR mecanum wheels used in this work. The system is equipped with two key subsystems:

1. Wheel Odometry (WO): This subsystem provides data on the robot's linear and angular velocities. It utilizes the Dead Reckoning method to estimate the robot's position over time.

2. Indoor Positioning System (IPS) Sensor: This sensor directly measures and provides the robot's X and Y coordinates.

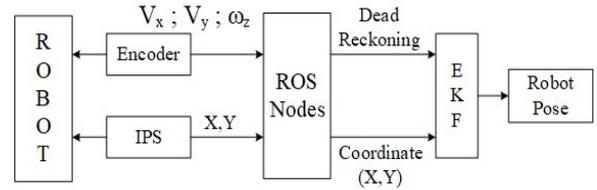

Fig. 1. System architecture of WMR.

The integration and processing of data from these components are managed by the Robot Operating System (ROS). ROS facilitates the concurrent reception and combination of sensor data through its receiving nodes.

To obtain an accurate estimation of the robot's pose, the system employs the EKF fusion algorithm. This algorithm fuses the position data from both the WO and IPS sensors, resulting in a more precise estimate of the robot's overall pose.

### B. Hardware Setup

In this work, we use a WMR platform equipped with 4 Mecanum wheels. Mecanum wheels are particularly advantageous for WMRs due to their ability to facilitate omnidirectional movement. . This capability allows the robot to move in any direction without requiring adjustments to the wheel alignment, making it highly maneuverable in constrained environments [12], [13].

The platform is equipped with a Raspberry Pi 4 running ROS. The L298D motor controller and four encoders with 1700 PPR are used to control four DC motors. The IPS sensor used in this work is a solution provided by Marvelmind Robotics. The mobile beacon is placed on the robot platform, while the beacon station is placed on a tripod with a 2-meter height in the corner of the experimental room. All sensor data, including encoder pulse and coordinate readings using the IPS sensor, are available in the ROS message server.

### C. Ultrasound based Indoor Positioning System

The Indoor Positioning System (IPS) consists of three main components: a modem, stationary beacons, and a mobile beacon. The modem serves as the central controller, managing communication among components and interfacing with dashboard software. Stationary beacons, fixed in position, emit and receive ultrasound signals to provide spatial reference points. The mobile beacon, mounted on the robot, interacts with these stationary beacons to determine its precise location within the indoor environment.

Fig. 3. illustrates the configuration of the IPS setup. We use the commercial Marvelmind Indoor Navigation System. Marvelmind provides a Robot Operating System (ROS)

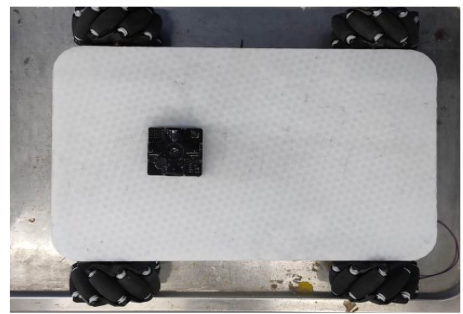

Fig. 2. WMR mecanum wheels.

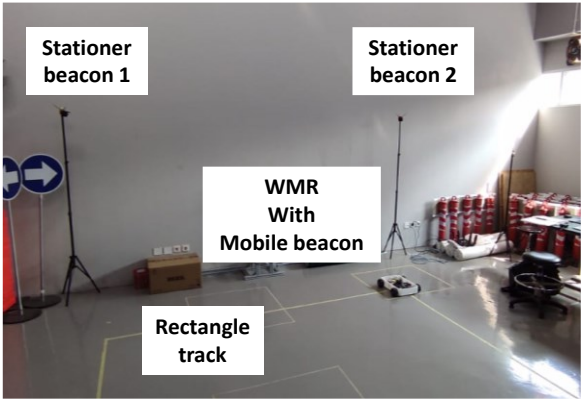

Fig. 3. IPS setup configurations.

package that facilitates communication with the mobile beacon and enables the retrieval of location data.

This setup ensures accurate tracking of the robotic vehicle's position by integrating the data from the stationary and mobile beacons through the Marvelmind system and ROS.

### D. Wheel Odometry-Based Localization for Mecanum-Wheeled Robots

Fig. 4 illustrates the model of the Mecanum-wheeled robot used in this study. The robot's kinematics can be mathematically represented by the kinematic equation:

$$\begin{bmatrix} V_x \\ V_y \\ \omega_z \end{bmatrix} = \frac{r}{4} \begin{bmatrix} 1 & 1 & 1 & 1 \\ -1 & 1 & 1 & -1 \\ -\frac{1}{(l_{fr}+l_{rl})} & \frac{1}{(l_{fr}+l_{rl})} & -\frac{1}{(l_{fr}+l_{rl})} & \frac{1}{(l_{fr}+l_{rl})} \end{bmatrix} \begin{bmatrix} \omega_1 \\ \omega_2 \\ \omega_3 \\ \omega_4 \end{bmatrix} \quad (1)$$

In which:
- $V_x$ represents the linear velocity along the x-axis, which corresponds to the forward and backward motion of the robot.
- $V_y$ denotes the linear velocity along the y-axis, which relates to the lateral (side-to-side) movement.
- $\omega_z$ is angular velocity around the z-axis, defining rotational movement of the robot about its vertical axis.

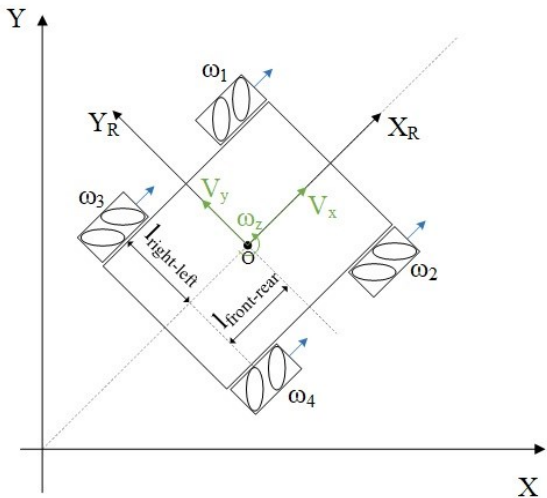

Fig. 4. Mobile robot kinematic with mecanum wheels.

- Positive values in $V_x$, $V_y$, and $\omega_z$ in sequences are in the forward, left, and anticlockwise directions.
- r (radius of the mecanum wheel) is 46.875 mm
- $l_{fr}$ (half distance between the front-rear wheels) is 135 mm
- $l_{rl}$ (half distance between right-left wheels) is 125 mm

The updated pose of the Mecanum-wheeled mobile robot can be determined using the dead reckoning formula, which is expressed as follows:

$$\begin{bmatrix} X_{k+1} \\ Y_{k+1} \\ \theta_{k+1} \end{bmatrix} = \begin{bmatrix} X_k \\ Y_k \\ \theta_k \end{bmatrix} + \begin{bmatrix} cos_{\theta k} & -sin_{\theta k} & 0 \\ sin_{\theta k} & cos_{\theta k} & 0 \\ 0 & 0 & 1 \end{bmatrix} \begin{bmatrix} V_x \\ V_y \\ \omega_z \end{bmatrix} dt \quad (2)$$

In this equation:
- $(X_{k+1}, Y_{k+1}, \theta_{k+1})$ represents the updated pose of the robot at the next time step, $t_{k+1}$, where X and Y are the Cartesian coordinates, and $\theta$ is the orientation angle (heading) of the robot.
- $(X_k, Y_k, \theta_k)$ denotes the robot's pose at the current time step, $t_k$.
- The matrix multiplication involving the rotation matrix accounts for the orientation of the robot at time $t_k$, aligning the velocity components and $V_y$ with the robot's frame of reference.
- $V_x$ and $V_y$ are the linear and angular velocities of the robot, respectively, as determined by the kinematic model.
- $dt = t_{k+1} - t_k$ is the time interval between the current and next pose estimations.

This formula provides a method for incrementally updating the robot's pose based on its current velocity and orientation, allowing for real-time tracking of its position and heading as it moves. The use of dead reckoning can accumulate errors over time, necessitating occasional corrections from external positioning systems or sensors to maintain accuracy.

### E. EKF based Fusion for State Estimation

#### 1) System Model

The kinematic model of a Mecanum-wheeled mobile robot differs from traditional wheeled robots due to the omnidirectional capabilities provided by the Mecanum wheels. This allows the robot to move in any direction regardless of its orientation. The system model for this robot includes the position, orientation, and velocities, considering the unique motion characteristics enabled by the Mecanum wheels.

##### a) State Vector:

The state vector $\mathbf{x}_k$ represents the pose of the robot, including its position and orientation:

$$\mathbf{x}_k = \begin{bmatrix} x_k \\ y_k \\ \theta_k \end{bmatrix} \quad (3)$$

where $x_k$ and $y_k$ are the Cartesian coordinates of the robot's position, and $\theta_k$ is the orientation angle (yaw) of the robot.

*b) Control Input Vector:*

The control input vector $\mathbf{u}_k$ represents the robot's velocities:

$$\mathbf{u}_k = \begin{bmatrix} v_x \\ v_y \\ \omega \end{bmatrix} \quad (4)$$

where $v_x$ and $v_y$ are the linear velocities along the x and y axes in the robot's frame of reference, and $\omega$ is the angular velocity (rate of change of orientation $\theta$).

*c) Process Model:*

The process model describes how the state evolves over time based on the control inputs. For a Mecanum wheeled robot, the motion model accounts for the ability to move laterally, forward, backward, and rotate:

$$\mathbf{x}_{k+1} = \mathbf{x}_k + \begin{bmatrix} \Delta t(v_x \cos\theta_k - v_y \sin\theta_k) \\ \Delta t(v_x \sin\theta_k + v_y \cos\theta_k) \\ \Delta t \omega \end{bmatrix} + \mathbf{w}_k \quad (5)$$

where $\Delta t$ is the time step between state updates, and $\mathbf{w}_k$ is the process noise vector, representing uncertainties in the robot's motion.

*d) Jacobian of the Process Model:*

The Jacobian matrix $\mathbf{F}_k$ linearizes the process model around the current state estimate and is used in the EKF to propagate the covariance matrix:

$$\mathbf{F}_k = \frac{\partial \mathbf{f}(x_k, u_k)}{\partial \mathbf{x}_k} = \begin{bmatrix} 1 & 0 & -\Delta t(v_x \sin\theta_k + v_y \cos\theta_k) \\ 0 & 1 & \Delta t(v_x \cos\theta_k - v_y \sin\theta_k) \\ 0 & 0 & 1 \end{bmatrix} \quad (6)$$

*e) Measurement Model*

The measurement model relates the state of the robot to the observations made by the sensor, specifically the odometry and ultrasound-based indoor positioning system (IPS) proposed in this paper. The measurement model involves the distances from known landmarks or obstacles:

$$\mathbf{z}_k = \mathbf{h}(\mathbf{x}_k) + \mathbf{v}_k \quad (7)$$

where $\mathbf{z}_k$ is the vector of sensor measurements and $\mathbf{v}_k$ is the measurement noise vector.

*f) Jacobian of the Measurement Model:*

The Jacobian $\mathbf{H}_k$ linearizes the measurement model around the current state estimate:

$$\mathbf{H}_k = \frac{\partial \mathbf{h}(\mathbf{x}_k)}{\partial \mathbf{x}_k} \quad (8)$$

This matrix depends on the specific nature of the measurements and how they relate to the state variables.

*g) Noise Covariance Matrices:*

The EKF uses two noise covariance matrices, process noise covariance matrix $\mathbf{Q}_k$ and measurement noise covariance matrix $\mathbf{R}_k$, to update the state estimate and its covariance, improving the accuracy of localization.

The process noise covariance matrix $\mathbf{Q}_k$ represents the uncertainty in the robot's motion model. It accounts for the errors due to wheel slippage, uneven surfaces, and other factors affecting the robot's movement. $\mathbf{Q}_k$ can be defined based on the expected variability in the control inputs and the noise characteristics of the motion model. For example, assuming the noise in linear velocity $v$ and angular velocity $\omega$ are independent and Gaussian, $\mathbf{Q}_k$ can be structured as:

$$\mathbf{Q}_k = \begin{bmatrix} \sigma_v^2 & 0 \\ 0 & \sigma_\omega^2 \end{bmatrix} \quad (9)$$

where $\sigma_v^2$ and $\sigma_\omega^2$ are the variances of the process noise in linear and angular velocities, respectively.

The measurement noise covariance matrix $\mathbf{R}_k$ represents the uncertainty in the IPS sensor measurements. It accounts for errors such as measurement noise, environmental interference, and sensor inaccuracies. The measurement noise covariance matrix $\mathbf{R}_k$ is typically defined based on the sensor specifications and the expected variability in the measurements. For an IPS sensor, $\mathbf{R}_k$ can be structured as:

$$\mathbf{R}_k = \begin{bmatrix} \sigma_{IPS}^2 & 0 \\ 0 & \sigma_{IPS}^2 \end{bmatrix} \quad (10)$$

where $\sigma_{IPS}^2$ represents the variance in the IPS sensor measurements.

The variances of the process noise for linear and angular velocities were determined through experiments, where deviations in robot movement were analyzed under controlled conditions. For the IPS sensor, the measurement noise variance was based on the manufacturer's specifications and verified through calibration tests.

*2) Extended Kalman Filter Equations*

The EKF is used to estimate the state of the mobile robot by iteratively performing the following steps:

*a) Prediction Step:*

$$\hat{\mathbf{x}}_{k|k-1} = \mathbf{f}(\hat{\mathbf{x}}_{k-1|k-1}, \mathbf{u}_k) \quad (11)$$

$$\mathbf{P}_{k|k-1} = \mathbf{F}_k \mathbf{P}_{k-1|k-1} \mathbf{F}_k^T + \mathbf{Q}_k \quad (12)$$

where $\mathbf{F}_k$ is the Jacobian of $\mathbf{f}$ with respect to $\mathbf{x}$.

*b) Update Step:*

$$\mathbf{y}_k = \mathbf{z}_k - \mathbf{h}(\hat{\mathbf{x}}_{k|k-1}) \quad (13)$$

$$\mathbf{S}_k = \mathbf{H}_k \mathbf{P}_{k|k-1} \mathbf{H}_k^T + \mathbf{R} \quad (14)$$

$$\mathbf{K}_k = \mathbf{P}_{k|k-1} \mathbf{H}_k^T \mathbf{S}_k^{-1} \quad (15)$$

$$\hat{\mathbf{x}}_{k|k} = \hat{\mathbf{x}}_{k|k-1} + \mathbf{K}_k \mathbf{y}_k \quad (16)$$

$$\mathbf{P}_{k|k} = (\mathbf{I} - \mathbf{K}_k \mathbf{H}_k) \mathbf{P}_{k|k-1} \quad (17)$$

where $\mathbf{H}_k$ is the Jacobian of $\mathbf{h}$ with respect to $\mathbf{x}$, and $\mathbf{K}_k$ is the Kalman gain.

## IV. EXPERIMENTAL SETUP

Fig. 5 shows the controlled environment designed to evaluate the robot's navigational capabilities and the effectiveness of the state estimation algorithm. The test area is a 3-meter by 3-meter rectangular space. It is clearly marked by tape lines on the floor, which define the boundaries of the experimental zone. The robot is remotely controlled to follow a square trajectory within the experimental area, which serves as a standardized path for testing its state estimation. The trajectory involves the following sequence of movements:

1. Initial Forward Movement (Positive x Direction): The robot begins at a designated starting point and moves forward along the x-axis (Point A).
2. Rightward Sliding (Negative y Direction): Upon reaching the edge of the rectangle, the robot controlled to perform a lateral movement to the right without altering its orientation (Point B).
3. Backward Movement (Negative x Direction): The robot then moves backward along the x-axis, returning along the opposite side of the rectangle (Point C).
4. Leftward Sliding (Positive y Direction): Finally, the robot slides left, returning to the starting point while maintaining its initial orientation (Point D). This completes the square trajectory and provides a comprehensive test of the robot's ability to accurately estimate its position.

The robot repeats this square loop trajectory multiple times to gather sufficient data for analysis. Repeating the motion allows for the assessment of the consistency and reliability of the robot's state estimation systems over time. It also provides insight into the cumulative error that may arise from dead reckoning and how effectively the EKF-based fusion corrects these errors.

## V. RESULTS AND DISCUSSION

Fig. 6 shows the WMR trajectory estimation using wheel odometry (a), indoor positioning system (b) and EKF based fusion (c) in comparison to the ground truth trajectory (black line). We aim the WMR movement is rectangle shape with 3-meter x 3-meter dimensions. Wheel speed different have an influence for changing the orientation of WMR. The wheel speed difference occurs because of the influence of terrain surface. The surface where experiment held, has uneven terrain. Due to wheel speed difference, the orientation of WMR is change incrementally, creating a slope in forward straight line and also in sliding movement. As a result, the trajectory of WMR is not a rectangle shape.

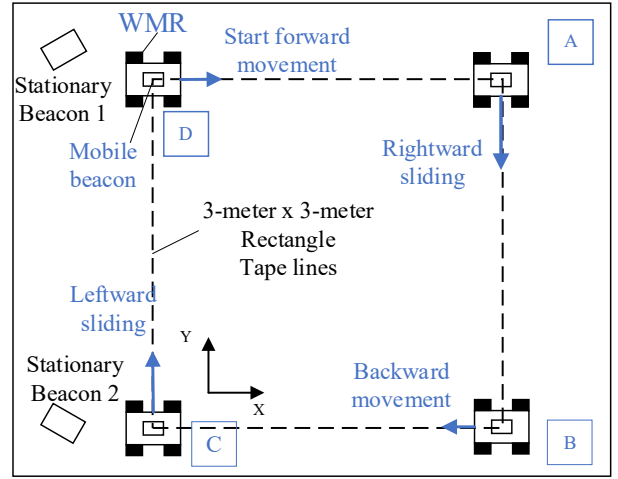

Fig. 5. Experimental area layout.

Result of wheel odometry is presented in blue line in Fig. 6a. It can be seen that during the first and second movement of the WMR, the wheel odometry trajectory closely follows the ground truth line, indicating satisfactory accuracy. However, as the robot progresses to point C and D, the trajectory begins to deviate due to angular error inherent in the single odometry dead reckoning approach. These errors accumulate, leading to a growing discrepancy in robot pose estimation. By the final pose, the wheel odometry trajectory has drifted significantly from the origin (0,0).

A key limitation of wheel odometry is its reliance on wheel encoder sensor reading, which are susceptible to errors such as wheel slippage. Wheel slippage occurs when there is a discrepancy between the robot's actual velocity and the velocity reported by the wheel encoder [14]. This discrepancy leads to errors in the forward and angular velocity calculations derived from the encoder readings, ultimately causing the drift observed in the odometry results.

The IPS measurement results, shown as the red line in Fig. 6b, indicate that the IPS trajectory closely aligns with the ground truth during the first and third movements of the WMR, demonstrating a reasonable level of accuracy. However, noticeable discrepancies appear during the second and fourth movements compared to the ground truth line. Despite these variations, the IPS measurements do not display the cumulative drift observed in wheel odometry results. A key limitation of the IPS sensor is the presence of noise in the positioning data, which, while causing fluctuations in the robot's estimated position, does not lead to the gradual drift typical of odometry-based methods.

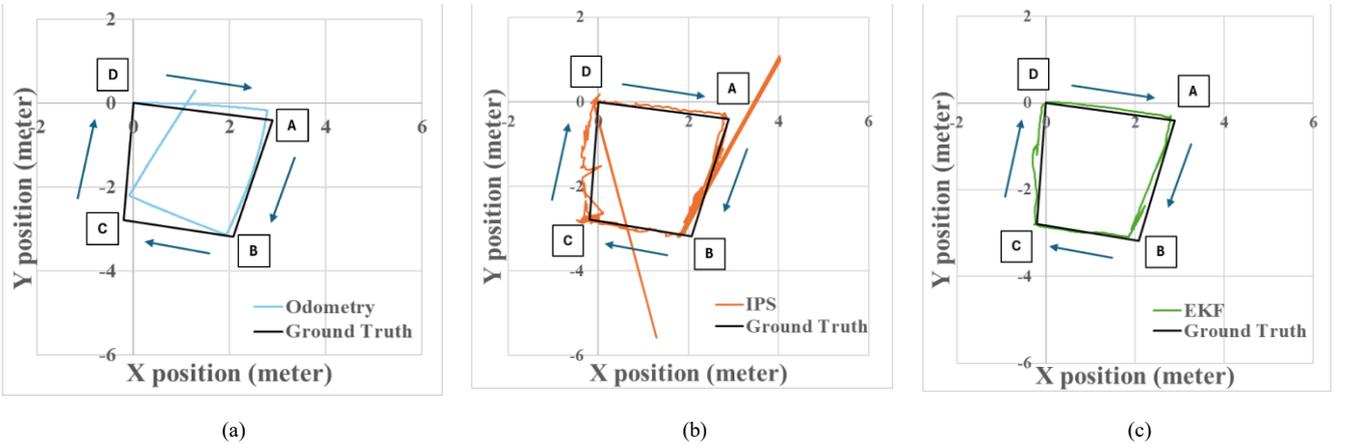

Fig. 6. Comparison of trajectories from (a) Wheel odometry, (b) IPS, and (c) EKF estimated with ground truth.

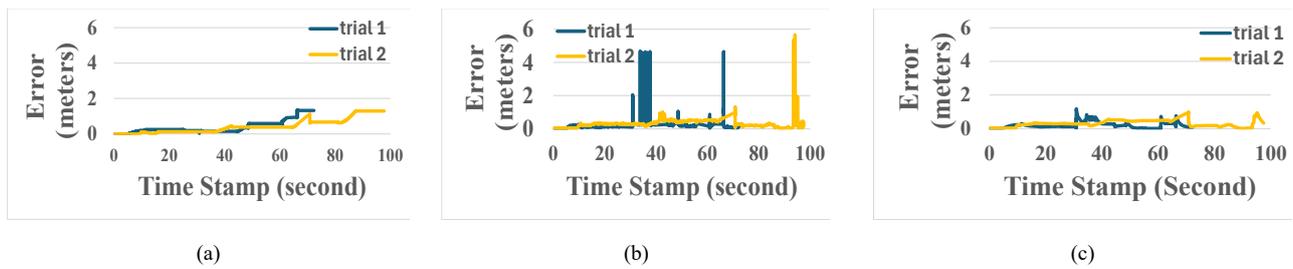

Fig. 7. Comparison of distance error from (a) Wheel odometry, (b) IPS, (c) EKF estimated with ground truth.

The EKF fusion results, shown as the green line in Fig. 6c, demonstrate a clear improvement over the wheel odometry-only approach, with a significant reduction in error. EKF fusion effectively mitigates errors inherent to wheel odometry, such as those caused by wheel slippage, while compensating for noise in the IPS positioning data by incorporating the more stable trajectory information from wheel odometry. This combination of position estimates from both wheel odometry and IPS measurements yields a more accurate and reliable overall trajectory.

Further analysis of the distance errors highlights these improvements. As shown in Fig. 7a, the wheel odometry-only approach results in errors peaking at 1.3 meters at points C and D, mainly due to angular inaccuracies. In contrast, Fig. 7b shows that IPS measurements, affected by coordinate estimation noise, produce a maximum error of 4.6 meters. The EKF estimation, presented in Fig. 7c, significantly enhances trajectory tracking by reducing angular errors from wheel odometry and compensating for IPS noise, resulting in a minimal error of 1.1 meters. This comparison demonstrates the EKF's effectiveness in enhancing localization accuracy by addressing the limitations of both wheel odometry and IPS.

## VI. Conclusion

This work proposes a comprehensive method for localizing mobile robots using sensor fusion techniques that integrate wheel odometry data with an ultrasound-based indoor positioning system (IPS). Firstly, hardware research platform for an indoor localization wheeled mobile robot is built in this study. Secondly, a kinematic study was carried out on wheeled mobile robot and its motion is used in wheel odometry derivation trajectory method. Furthermore, the IPS sensor is installed in wheeled mobile robot, where it directly provides the robot position in the X and Y coordinates. In contrast, the wheel odometry uses the Dead Reckoning method to provide these coordinates. The key contribution of this work lies in the fusion of IPS location data with wheel odometry data using the Extended Kalman Filter (EKF) method. This fusion significantly enhance the localization accuracy by combining the strengths of both techniques—leveraging the precision of IPS and the continuous tracking capability of wheel odometry—while mitigating the limitations inherent in each method when used independently.